\documentclass[letterpaper]{article} 
\usepackage{aaai2026} 
\usepackage{times}  
\usepackage{helvet}  
\usepackage{courier} 
\usepackage[hyphens]{url} 
\usepackage{graphicx} 
\usepackage{natbib}  
\usepackage{caption} 
\usepackage{algorithm}
\usepackage{algorithmic}
\usepackage{amsmath,amsfonts}
\usepackage{booktabs}
\usepackage{multirow}
\usepackage{makecell}
\usepackage{xcolor} 
\usepackage{subcaption}

\usepackage{cite} 
\usepackage{newfloat}
\usepackage{listings}
\DeclareCaptionStyle{ruled}{labelfont=normalfont,labelsep=colon,strut=off} 
\floatstyle{ruled}
\newfloat{listing}{tb}{lst}{}
\floatname{listing}{Listing}

\title{Multi-Granularity Feature Calibration via VFM for \\Domain Generalized Semantic Segmentation}
\author {
    Xinhui Li\textsuperscript{\rm },
    Xiaojie Guo\textsuperscript{\rm }\thanks{Corresponding Author},
}
\affiliations {
    \textsuperscript{\rm }Tianjin University, Tianjin, China\\
    lixinhui@tju.edu.cn, xj.max.guo@gmail.com 
}

\usepackage{bibentry}

\begin{document}

\maketitle

\begin{abstract}
Domain Generalized Semantic Segmentation (DGSS) aims to improve the generalization ability of models across unseen domains without access to target data during training. Recent advances in DGSS have increasingly exploited vision foundation models (VFMs) via parameter-efficient fine-tuning strategies. However, most existing approaches concentrate on global feature fine-tuning, while overlooking hierarchical adaptation across feature levels, which is crucial for precise dense prediction. In this paper, we propose Multi-Granularity Feature Calibration (MGFC), a novel framework that performs coarse-to-fine alignment of VFM features to enhance robustness under domain shifts. Specifically, MGFC first calibrates coarse-grained features to capture global contextual semantics and scene-level structure. Then, it refines medium-grained features by promoting category-level feature discriminability. Finally, fine-grained features are calibrated through high-frequency spatial detail enhancement. By performing hierarchical and granularity-aware calibration, MGFC effectively transfers the generalization strengths of VFMs to the domain-specific task of DGSS. Extensive experiments on benchmark datasets demonstrate that our method outperforms state-of-the-art DGSS approaches, highlighting the effectiveness of multi-granularity adaptation for the semantic segmentation task of domain generalization.
\end{abstract}

\section{Introduction}
Benefiting from large-scale supervised data, deep neural networks have achieved remarkable performance in a wide range of computer vision tasks, such as object detection~\cite{he2017mask,cheng2018learning} and semantic segmentation~\cite{strudel2021segmenter,ding2020semantic}. 
A core premise behind these advances is the assumption that training and test data are independently and identically distributed.
However, in realistic scenarios, this assumption often breaks down due to unpredictable distribution shifts, especially since test data is typically unknown and inaccessible during training. To address these challenges, domain generalized semantic segmentation (DGSS) has emerged as a crucial research direction~\cite{Yue2019DomainRA,Matsuura2020DomainGU,Li2021SemanticSW,wu2022siamdoge}. DGSS aims to train segmentation models solely on labeled source domains while ensuring robust generalization performance across diverse and unseen target domains during inference.

\begin{figure}
\centering
\includegraphics[width=1.0\columnwidth]{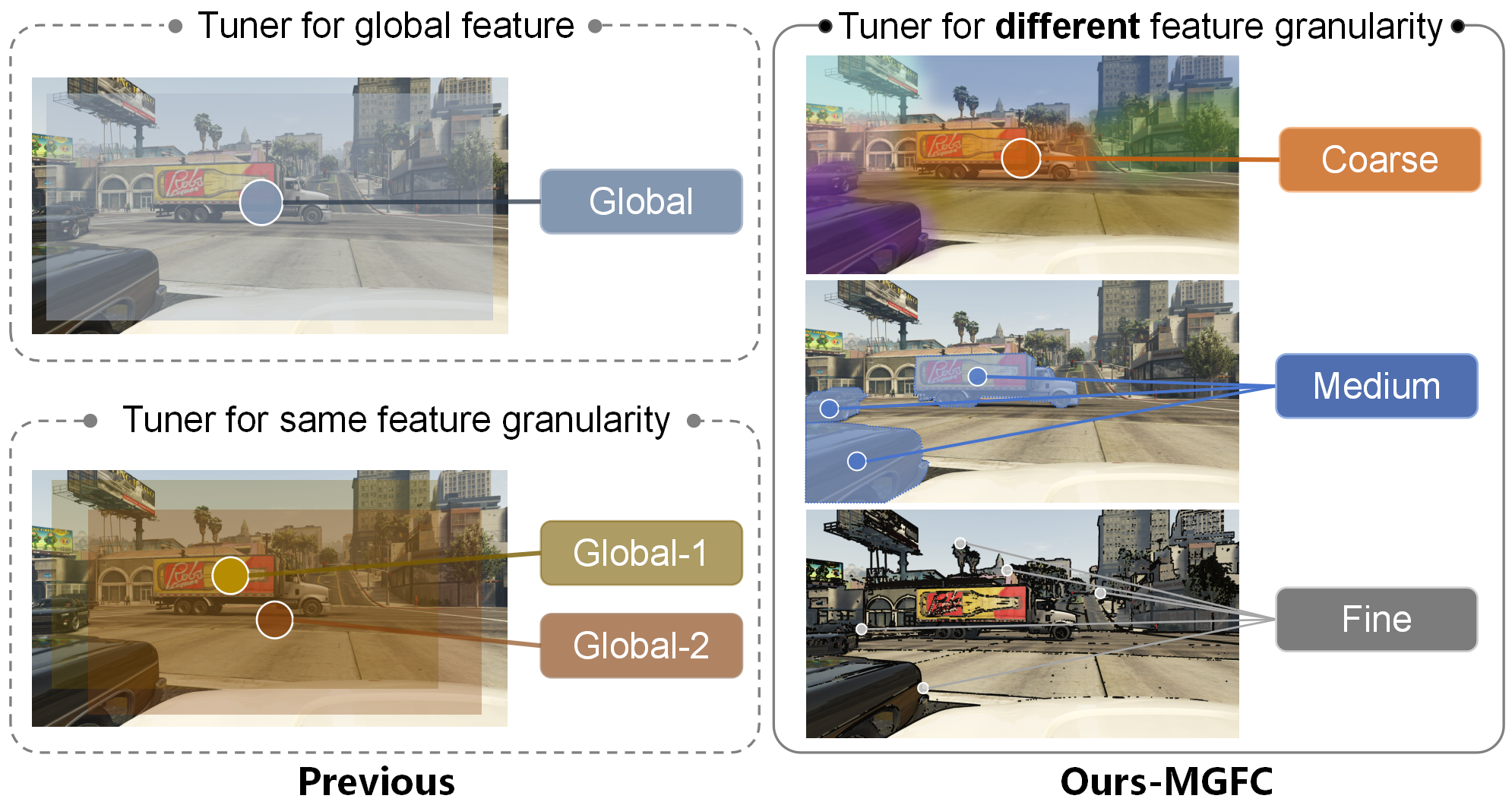} 
\caption{Illustration of the existing DGSS methods and Ours. \textbf{Left:} Previous methods focus on feature fine-tuning at the same granularity. \textbf{Right:} Our method achieves \emph{multi-granularity feature calibration} at different feature levels.}
\label{intro}
\end{figure}

Recent advancements in DGSS researches ~\cite{wei2024stronger,yi2024learning,bi2024learning} have been significantly influenced by the emergence of Vision Foundation Models (VFMs), which offer substantial advantages over earlier methods~\cite{Li2018DomainGW,Peng2019MomentMF,Matsuura2020DomainGU,Zhu2021SelfsupervisedUD}. 
Pretrained on massive and diverse datasets spanning various modalities and tasks, VFMs such as DINO~\cite{oquab2023dinov2}, SAM~\cite{kirillov2023sam}, and CLIP~\cite{radford2021learning} demonstrate impressive zero-shot and few-shot generalization capabilities. 
These models inherently capture rich visual semantics and exhibit strong robustness to distribution shifts. 
Notably, recent work~\cite{wei2024stronger} shows that even frozen VFMs, when directly applied to DGSS without any task-specific fine-tuning, can outperform many existing domain generalization methods. This finding underscores the unique generalization potential embedded in VFMs and reveals their capacity to serve as strong inductive priors for dense prediction tasks under domain shift. 

In parallel, the emergence of parameter-efficient fine-tuning techniques, such as Low-Rank Adaptation (LoRA)~\cite{hu2021lora}, has further enhanced the application of VFMs in DGSS. Recent methods combining frozen VFMs with modular fine-tuning~\cite{wei2024stronger,yi2024learning,bi2024learning} have achieved notable success. These approaches leverage pre-trained VFMs to achieve domain generalization while requiring minimal updates to their network parameters. However, current methods focus primarily on global feature adaptation, even when decomposing feature representations~\cite{yi2024learning,bi2024learning} within specific layers. As shown in Fig.~\ref{intro}, most existing methods, whether based on fine-tuning or multi-branch adaptation, primarily operate at a global level.
While global adaptation plays an important role in DGSS, fine-grained calibration is equally essential for achieving robust performance.
In particular, pixel-wise segmentation accuracy is highly dependent on object-level semantic consistency and precise boundary delineation, which are often overlooked by global alignment strategies.
This raises a critical question: \textit{How can fine-tuning methodologies address the granularity required for the pixel-level prediction task in DGSS?}

To tackle this challenge, we propose a novel \textbf{Multi-Granularity Feature Calibration} (MGFC) framework for DGSS. As shown in Fig.~\ref{intro}, MGFC performs hierarchical feature alignment across three granularity levels enabling comprehensive fine-tuning from global semantics to local details. Specifically, \textit{Coarse-grained calibration} preserves global contextual representations and captures scene-specific semantics critical for global feature alignment.
\textit{Medium-grained calibration} guides the generalization power of VFMs to concentrate on category-level semantics, enhancing task-specific feature representation.
\textit{Fine-grained calibration} emphasizes spatial refinement by reinforcing high-frequency details, facilitating precise edge localization and pixel-level segmentation accuracy.
By explicitly modeling feature adaptation across multiple granularities, MGFC bridges the gap between global robustness and local precision.
To further integrate multi-granularity representations, we propose a query fusion module that combines contextual global patterns with precise local details. This module enhances performance on downstream DGSS tasks while preserving the generalization capacity of the frozen VFMs. Overall, the major contributions of this work can be summarized as follows:
\begin{itemize}
    \item We propose a novel multi-granularity feature calibration framework for DGSS, which enables hierarchical fine-tuning of vision foundation model features across coarse, medium, and fine levels.
    \item A query fusion module is desgined to effectively integrate multi-granularity representations, guiding the network to jointly capture both global semantics and fine-grained structural cues.
    \item 
    Extensive experiments and ablation studies demonstrate the effectiveness and superior performance of our method over state-of-the-art approaches.
\end{itemize}

\section{Related Work}
\subsection{Domain Generalization}
To address the challenge of performance degradation when models encounter data from unseen domains, domain generalization (DG) has emerged as a key strategy. DG methods~\cite{su2022consistency,li2022cross} are fundamentally designed to enhance the generalization of models under diverse or unknown distribution conditions. 
Specifically, DG concentrates on enhancing generalization in entirely unseen domains through multi-source training~\cite{gan2016learning, zhao2021learning}, feature normalization~\cite{lee2023decompose,qi2024normaug}, and data augmentation~\cite{su2023rethinking, wang2024inter,zhou2024mixstyle}. The aforementioned DG methods have been extensively applied to various downstream tasks, such as image classification tasks~\cite{zhou2021domain,Kang2022CVPR,Zhang2022ExactFD} and semantic segmentation~\cite{Yue2019DomainRA,choi2021robustnet,wu2022siamdoge}. Most DG methods~\cite{li2022cross,jiang2023domain, Fahes_2024_CVPR} are built on classical network backbones, such as ResNet~\cite{he2016deep}, VGGNet~\cite{simonyan2014very}, and ShuffleNetV2~\cite{ma2018shufflenet}.
However, with the advent of vision foundation models(VFM)~\cite{he2022mae,radford2021clip,kirillov2023sam,fang2023eva02,fang2023eva,oquab2023dinov2}, recent approaches in DG have increasingly focused on exploring how to adapt the generalization capabilities of VFMs to improve model's performance.

\subsection{VFM-based DGSS}
Most current VFM-based DGSS methods~\cite{wei2024stronger,bi2024learning} adopt frozen VFMs as the backbone network, fine-tuning them to effectively utilize the generalization capabilities of VFM while minimizing training costs and time.
Specifically, the Rein framework~\cite{wei2024stronger} pioneered this direction by introducing a parameter-efficient fine-tuning approach. Building upon this foundation, several advanced approaches have emerged. The SET method~\cite{yi2024learning} addresses DGSS by decomposing frozen VFM features into phase and amplitude components in the frequency space. Meanwhile, FADA~\cite{bi2024learning} decouples domain-invariant content from domain-specific styles using the Haar wavelet transform. In addition, ~\cite{pak2024textual} leverages domain-invariant semantic knowledge from text embeddings of VFMs to enhance semantic clarity in dense visual features.
However, most existing approaches predominantly focus on global feature adaptation, overlooking the multi-granularity nature of semantic segmentation tasks, as shown in Fig.~\ref{intro}. DGSS requires accurate classification at various levels, from global scene understanding to fine-grained boundary delineation. To address this limitation, our work introduces a novel multi-granularity feature calibration network that fine-tunes VFM features at different levels of granularity.

\begin{figure*}[t]
\centering
\includegraphics[width=1.0\textwidth]{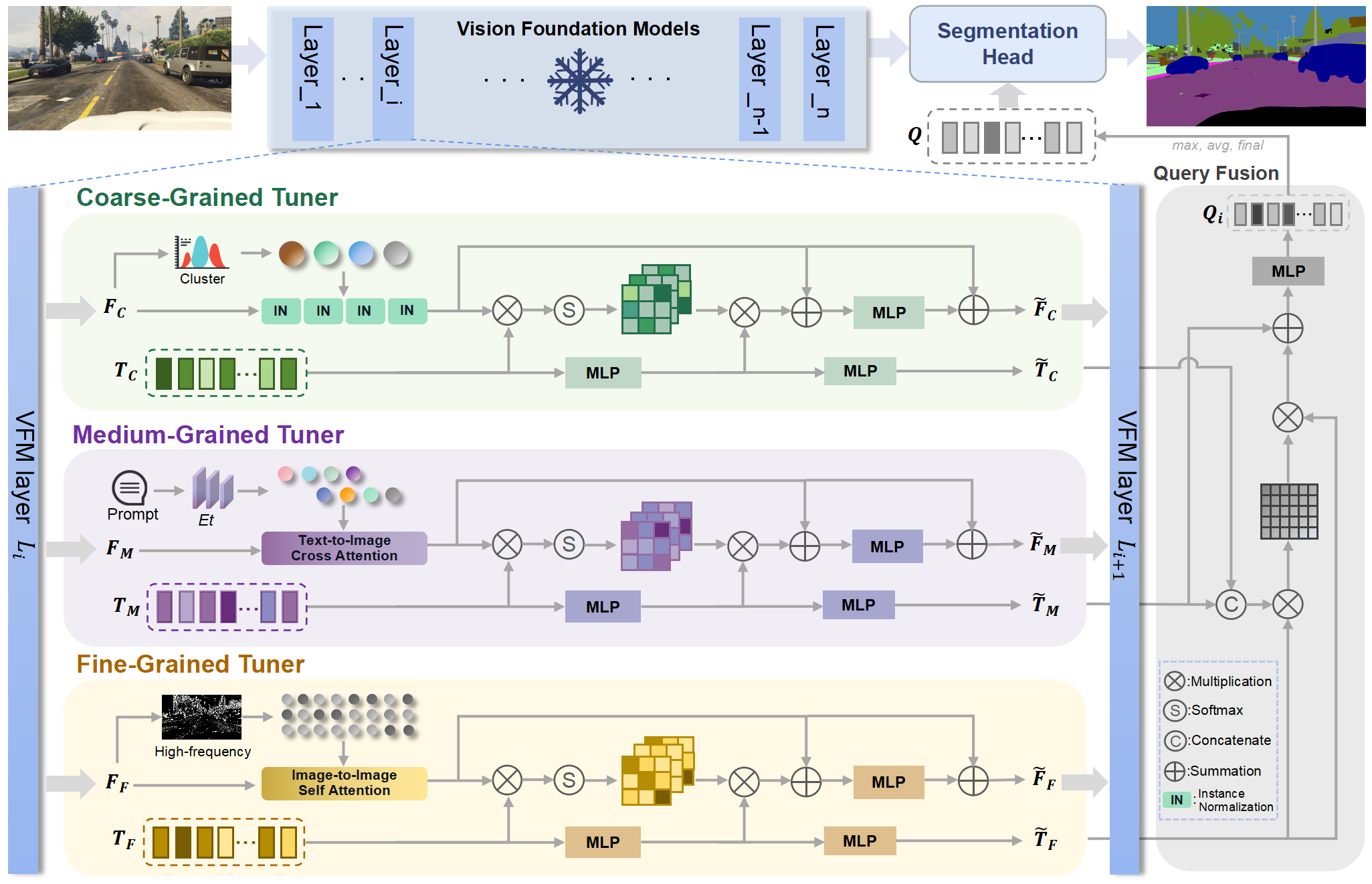} 
\caption{The network architecture of the multi-granularity feature calibration (MGFC) framework, including the coarse-grained tuner, medium-grained tuner, fine-grained tuner and the query fusion module.}
\label{network}
\end{figure*}

\section{Methodology}
\subsection{Overall Architecture}
The overall architecture of our method is illustrated in Fig.~\ref{network}. Given an input image, we employ VFM as the backbone with \( N \) sequential layers (denoted as \( L_1, L_2, \dots, L_N \)), and each layer corresponds to a multi-granularity feature calibration strategy.
To effectively adapt the frozen VFM features for domain generalized segmentation, we propose a hierarchical fine-tuning mechanism consisting of a \textit{coarse-grained tuner}, a \textit{medium-grained tuner}, a \textit{fine-grained tuner}, and a \textit{query fusion module}.

\subsection{Coarse-grained Tuner}
In the DGSS, it is crucial to reduce style variation and maintain a stable representation of scene content to predict scene semantics. Instance normalization~\cite{huang2019iterative,Pan2018TwoAO}, which standardizes each channel based on its mean and variance, is effective in suppressing style-related variations.
To this end, we design the Coarse-grained Tuner (CGT) to suppress domain-specific style variations from a global perspective through feature clustering and instance normalization.
Specifically, given a feature map \(\mathbf{F}_i \) extracted from each layer \( L_i \) of the VFM. In the CGT, we denote the \( \mathbf{F}_i \) as \( \mathbf{F}_{C_i} \). We first apply clustering on the \( \mathbf{F}_{C_i} \) along the spatial dimension and group them into \( K \) clusters. Let the resulting clusters be denoted as \( \ \mathbf{F}_{C_i}^{k}\), for \( k = 1, ... ,K \). Each cluster captures features with similar style or semantic patterns.
It is worth noting that we adopt different clustering strategies in this step, including DBSCAN and K-Means, and conduct a comparative analysis of their effectiveness in the experiment section.
Next, we perform instance normalization independently on each \( \mathbf{F}_{C_i}^{k} \) to remove domain-specific statistics while preserving semantic structure within the cluster. The normalized cluster features \(  \mathbf{\hat F}_{C_i}^{k} \) are obtained as: \(\mathbf{\hat F}_{C_i}^{k} = \operatorname{IN}(\mathbf{F}_{C_i}^{k})\), where \( \operatorname{IN}(\cdot) \) denotes instance normalization.
Finally, the normalized cluster features \( \mathbf{\hat F}_{C_i}^{k}\) are concatenated along the spatial dimension to reconstruct the full feature map \( \mathbf{\hat F}_{C_i} \), which has the same dimension as the original feature \( \mathbf{F}_{C_i} \). This coarse-grained clustering not only enables global style suppression but also mitigates the risk of over-normalization that may arise from applying instance normalization directly to the entire feature map, thereby preserving essential structural and semantic information.

Meanwhile, motivated by Rein~\cite{wei2024stronger}, which leverages learnable tokens to bridge feature representations and instances, we adopt a similar design to enable effective feature fine-tuning from the VFM to downstream tasks. As illustrated in the CGT branch of Fig.~\ref{network}, \( \mathbf{T}_{C_i} \in \mathbb{R}^{m \times c} \) denote the learnable coarse-grained token, where \( m \) is the sequence length and \( c \) is the channel dimension of the \( L_i \). The token \( \mathbf{T}_{C_i} \) serves to capture coarse-grained scene content from \( \mathbf{\hat F}_{C_i} \), while adhering to the low-rank adaptation paradigm.
Firstly, the similarity map \( \mathbf{S}_{C_i}\) is calculated between the token \( \mathbf{T}_{C_i} \) and feature \( \mathbf{\hat F}_{C_i} \), which measures the correlation between token embeddings and the patch embeddings of the feature map. The similarity map \( \mathbf{S}_{C_i}\) is computed as:
\(
    \mathbf{S}_{C_i} = \operatorname{Softmax} \left( \frac{\mathbf{\hat F}_{C_i} \times {\mathbf{T}_{C_i}^{\mathrm{T}}}
}{\sqrt{c}} \right),
\)
where \(\operatorname{Softmax}\) represents the softmax activation function. 
Subsequently, the \( \mathbf{T}_{C_i} \) is projected into the feature space of \( \mathbf{\hat F}_{C_i} \) using a multi-layer perceptron (MLP) parameterized by weights \( \mathbf{W}_{C_i}^{1} \) and bias \( \mathbf{b}_{C_i}^{1} \), followed by an alignment with the similarity map \( \mathbf{S}_{C_i} \). Then the projected feature is fused with the normalized features \( \mathbf{\hat F}_{C_i} \) using another MLP with weight parameters \( \mathbf{W}_{C_i}^{2} \) and bias parameters \( \mathbf{b}_{C_i}^{2} \) by a skip connection. This alignment enables the token to focus on the semantically relevant scene regions. The process can be expressed as:
\begin{align}
    \mathbf{\tilde{F}}_{C_i} = \mathbf{\hat F}_{C_i} &+ \Bigl( 
        \mathbf{S}_{C_i} \times \left( \mathbf{T}_{C_i} \times \mathbf{W}_{C_i}^{1} + \mathbf{b}_{C_i}^{1} \right) 
        \nonumber \\
        &+ \mathbf{\hat F}_{C_i}  \Bigr) \times \mathbf{W}_{C_i}^{2} + \mathbf{b}_{C_i}^{2},
\end{align}
where \( \mathbf{\tilde{F}}_{C_i} \) denotes the final enhanced coarse-grained features used for the next layer \( L_{i+1} \).

\subsection{Medium-grained Tuner}
In the Medium-grained Tuner (MGT), we leverage object-level cues to refine visual features, encouraging the frozen VFM representations to focus on task-specific semantics.
We employ CLIP’s pre-trained text encoder~\cite{radford2021learning} to incorporate object-level semantic priors. Specifically, we input category texts such as \textit{``car''}, \textit{``road''}, \textit{``traffic sign''}, etc., as textual prompts. 
Given a set of object-level textual descriptions \( \{t_1, t_2, \dots, t_n\} \), we use the pre-trained text encoder \( E_T \) to extract their semantic embeddings as:  
\(
\mathbf{F}_T = E_T\left( \{t_1, t_2, \dots, t_n\} \right) \in \mathbb{R}^{n \times c},
\)
where \( n \) denotes the number of object categories and \( c \) is the feature dimensionality. \( \mathbf{F}_T \) are then integrated with the intermediate visual features \( \mathbf{F}_{M_i} \) extracted from the VFM through a text-to-image cross-attention module. This fusion injects object-level semantic priors into the visual representations, facilitating more precise and semantically informed feature alignment. 
In the cross-attention module, the visual features \( \mathbf{F}_{M_i} \) are used as queries, while the textual embeddings \( \mathbf{F}_T \) serve as both keys and values. The cross-attended features \( \mathbf{\hat F}_{M_i} \) are computed as:
\begin{align}
    \mathbf{\hat F}_{M_i} &= \operatorname{CrossAttn}(Q = \mathbf{F}_{M_i}, K = \mathbf{F}_T, V = \mathbf{F}_T) \nonumber\\
    &= \operatorname{Softmax}\left(\frac{\mathbf{F}_{M_i} \times \mathbf{F}_T^\mathrm{T}}{\sqrt{d_k}}\right) \times\mathbf{F}_T,
\end{align}
where \( d_k \) denotes the dimensionality of the key embeddings. \( \operatorname{CrossAttn}(\cdot) \) denotes a multi-head attention operation. \( \operatorname{Softmax}\) represents the softmax activation function. The resulting features \( \mathbf{\hat F}_{M_i} \) integrate object-level semantic priors into the visual representations.
Next, to extract medium-grained feature representations from \( \mathbf{\hat{F}}_{M_i} \), we employ a learnable token \( \mathbf{T}_{M_i} \), following the low-rank adaptation paradigm. 
Specifically, we compute a similarity map \( \mathbf{S}_{M_i} \) between \( \mathbf{T}_{M_i} \) and \( \mathbf{\hat F}_{M_i} \), defined as:
\(
 \mathbf{S}_{M_i} = \operatorname{Softmax} \left( \frac{\mathbf{\hat F}_{M_i} \times \mathbf{T}_{M_i}^{\mathrm{T}}}{\sqrt{c}} \right),
\)
where \( c \) represents the channel dimension of the features and the token. 
Following a similar procedure as in the CGT, the mid-level token \( \mathbf{T}_{M_i} \) is aligned with the normalized feature \( \mathbf{\hat F}_{M_i} \) via a similarity map \( \mathbf{S}_{M_i} \), and enhanced through two MLPs with a skip connection. 
\( \mathbf{W}_{M_i}^{1}, \mathbf{W}_{M_i}^{2}, \mathbf{b}_{M_i}^{1}, \mathbf{b}_{M_i}^{2} \) denote two learnable weight matrices and their corresponding biases used for token projection and feature fusion, respectively.
The process is defined as:
\begin{align}
    \mathbf{\tilde{F}}_{M_i} = \mathbf{\hat F}_{M_i} &+ \Bigl( 
        \mathbf{S}_{M_i} \times \left( \mathbf{T}_{M_i} \times \mathbf{W}_{M_i}^{1} + \mathbf{b}_{M_i}^{1} \right)  \nonumber \\
        &+ \mathbf{\hat F}_{M_i}  \Bigr)
        \times \mathbf{W}_{M_i}^{2} + \mathbf{b}_{M_i}^{2}.
\end{align}

Here, \( \mathbf{\tilde{F}}_{M_i} \) denotes the mid-granularity features enriched by text-guided attention. 
Therefore, the refinement in MGT guide the model to attend more effectively to object-relevant semantic regions, facilitating the adaptation of VFMs to the domain-generalized segmentation task.

\subsection{Fine-grained Tuner} 
The ability to capture structural and fine-grained object details is essential for accurate pixel-level classification in DGSS. To encourage the VFMs to attend more to structural cues, we propose a Fine-grained Tuner (FGT) that enhances the structural representation through an image-to-image self-attention mechanism guided by high-frequency information. 
Specifically, we first extract the high-frequency components of the feature map \( \mathbf{F}_{F_i} \) from the \( i \)-th VFM layer using the Sobel operator, which computes gradients in both horizontal and vertical directions to emphasize edge and contour information. 
Such information is critical for structural understanding. The resulting high-frequency feature representation is used as the query in the self-attention module, while the original feature serves as both the key and the value. The structure-aware feature \( \mathbf{\hat{F}}_{F_i} \) is then formulated as:
\begin{align}
    \mathbf{\hat{F}}_{F_i} &= \operatorname{SelfAttn}\Bigl( Q = \operatorname{Sobel}(\mathbf{F}_{F_i}), \, K = \mathbf{F}_{F_i}, \, V = \mathbf{F}_{F_i} \Bigr) \nonumber\\
    &= \operatorname{Softmax}\left(\frac{\operatorname{Sobel}(\mathbf{F}_{F_i}) \times \mathbf{F}_{F_i}^\mathrm{T}}{\sqrt{d_k}}\right) \times \mathbf{F}_{F_i},
\end{align}
where \(\operatorname{Sobel}\) denotes the Sobel operator applied to each channel independently, and \(\operatorname{SelfAttn} (\cdot) \) refers to the multi-head self-attention mechanism. This strategy enables the model to strengthen feature responses along important structures, such as object boundaries and part contours, thereby enhancing fine-grained representations crucial for robust domain-invariant segmentation.

\begin{figure*}[t]
  \centering
\includegraphics[width=1.00\textwidth]{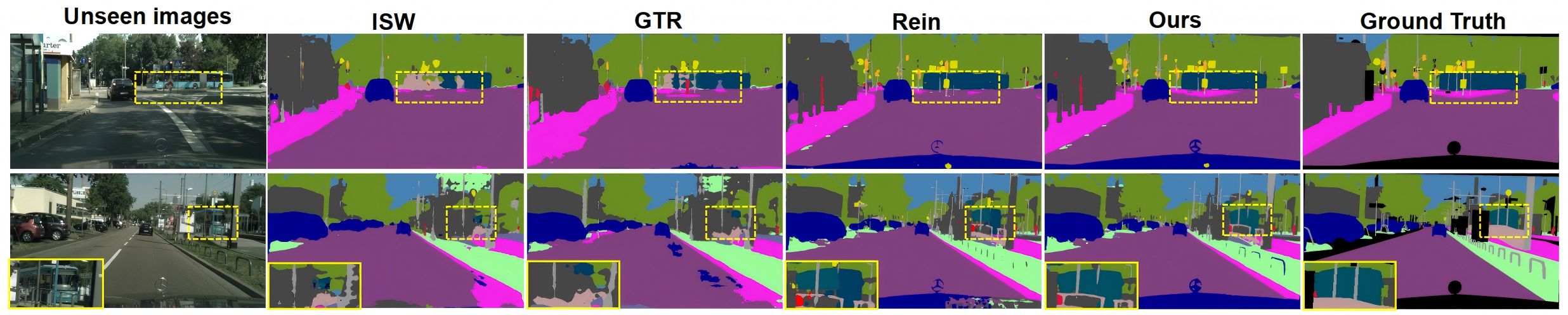}
   \caption{Qualitative results on domain generalized semantic segmentation. The model is trained on GTA5 and then generalized to Cityscapes using DINOv2.}
   \label{result-city}
   \vspace{-5pt}
\end{figure*}

In addition, similar to CGT and MGT, we also employ learnable fine-grained tokens \( \mathbf{T}_{F_i}\) in the FGT to guide the refinement of structural feature representations. The refined feature representation, denoted as \( \mathbf{\tilde{F}}_{F_i} \), is obtained by aligning the token with the structure-aware feature \( \mathbf{\hat{F}}_{F_i} \) through a similarity map \( \mathbf{S}_{F_i} \).
We compute the similarity map \( \mathbf{S}_{F_i} \) as:
\(
    \mathbf{S}_{F_i} = \operatorname{Softmax} \left( \frac{\mathbf{\hat{F}}_{F_i} \times \mathbf{T}_{F_i}^{\mathrm{T}}}{\sqrt{c}} \right),
\)
where \( c \) denotes the channel dimension. Then, the \( \mathbf{T}_{F_i} \) is then projected into the feature space via a learnable weight matrix \( \mathbf{W}_{F_i}^{1}\) and bias \( \mathbf{b}_{F_i}^{1}\), followed by a similarity-guided interaction and refinement:
\begin{align}
    \mathbf{\tilde{F}}_{F_i} = \mathbf{\hat{F}}_{F_i} &+ \Bigl( 
        \mathbf{S}_{F_i} \times \left( \mathbf{T}_{F_i} \times \mathbf{W}_{F_i}^{1} + \mathbf{b}_{F_i}^{1} \right) 
        \nonumber \\
        &+ \mathbf{\hat{F}}_{F_i} \Bigr) \times \mathbf{W}_{F_i}^{2} + \mathbf{b}_{F_i}^{2},
\end{align}
where \( \mathbf{W}_{F_i}^{2} \) and \( \mathbf{b}_{F_i}^{2} \) are the parameters of the second projection. This process guides the model to focus on fine-grained structural cues, thereby enhancing the structural detail representation passed to the subsequent layer \( L_{i+1} \).

Finally, for the entire architecture, the frozen VFM
feature \( \mathbf{F}_i \) at each layer \( L_i \) is passed through CGT, MGT, and FGT, respectively, yielding the features \(\mathbf{\tilde{F}}_{C_i}\), \(\mathbf{\tilde{F}}_{M_i}\), and \(\mathbf{\tilde{F}}_{F_i} \).
To achieve efficient feature fusion after each layer, we concatenate the features and apply a linear projection to align the dimensionality, as follows: \(\mathbf{\tilde{F}}_i=\operatorname{Linear} \left( \operatorname{Concat} \left( \mathbf{\tilde{F}}_{C_i}, \mathbf{\tilde{F}}_{M_i}, \mathbf{\tilde{F}}_{F_i} \right) \right)\), where the \(\operatorname{Linear}\) operation represents a linear transformation with a dimensionality of 1024. As a result, the projected feature \( \mathbf{\tilde{F}}_i \) is subsequently fed into the next VFM layer \( L_{i+1} \).

\subsection{Query Fusion Module}
At the core of the multi-granularity feature calibration strategy, three types of tokens are employed at different levels of granularity to guide the fine-tuning of VFM. Inspired by~\cite{wei2024stronger}, which establishes connections between learnable tokens and feature representations, we adopt a similar approach by linking the proposed tokens to the segmentation head. Before the connection, we design a query fusion module to fuse the coarse-grained token \( \mathbf{T}_{C_i} \), the medium-grained token \( \mathbf{T}_{M_i} \), and the fine-grained token \(\mathbf{T}_{F_i} \) to better capture multi-granularity feature presentations.

The fusion process starts by concatenating the coarse-grained token \( \mathbf{\tilde{T}}_{C_i} \) and the mid-grained token \( \mathbf{\tilde{T}}_{M_i} \) to form a joint representation: \(
\mathbf{\tilde{T}}_{CM_i} = \operatorname{Concat}(\mathbf{\tilde{T}}_{C_i}, \mathbf{\tilde{T}}_{M_i}),
\) which captures a broad range of global contextual cues. This fused token \( \mathbf{\tilde{T}}_{{CM_i}} \) then interacts with the fine-grained token \( \mathbf{\tilde{T}}_{F_i} \) through a cross-attention mechanism. The final multi-granularity representation is computed as:
\begingroup
\small
\begin{align}
    \mathbf{T}_{fuse_i} &= \mathbf{\tilde{T}}_{C_i} + \operatorname{CrossAttn}(Q = \mathbf{\tilde{T}}_{CM_i},\, K = \mathbf{\tilde{T}}_{F_i},\, V = \mathbf{\tilde{T}}_{F_i}) \nonumber \\
    &= \mathbf{\tilde{T}}_{C_i} + \operatorname{Softmax}\left( \frac{\mathbf{\tilde{T}}_{CM_i} \times \mathbf{\tilde{T}}_{F_i}^\mathrm{T}}{\sqrt{d_t}} \right) \times \mathbf{\tilde{T}}_{F_i},
\end{align}
\endgroup
where \( d_t \) is the dimensionality of the key embeddings and \( \operatorname{CrossAttn}(\cdot) \) is the multi-head attention operation. After the fusion process, we map the fused tokens \( T_{fuse_i}\) onto query \( \mathbf{Q}_i \) with a multi-layer perceptron.

For the queries from different layers, we adopt the simplified computation strategy proposed in ~\cite{wei2024stronger}, where the final query representation \( \mathbf{Q}_i \) is derived by aggregating the maximum and average components across all layer-wise queries. Specifically, for a set of queries \( \{\mathbf{Q}_i\}_{i=1}^{N} \), where N is the number of VMF
layers. We compute:
\(
\mathbf{Q}_{\text{max}} = \max_{i=1,2,\ldots,N} \mathbf{Q}_i,  \mathbf{Q}_{\text{avg}} = \frac{1}{N} \sum_{i=1}^N \mathbf{Q}_i,
\)
and the final query \( \mathbf{Q} \) entered into the segmentation header is obtained as: \( \mathbf{Q} = \text{Concat}([\mathbf{Q}_{\text{max}}, \mathbf{Q}_{\text{avg}}, \mathbf{Q}_N]) \times \mathbf{W}_Q + \mathbf{b}_Q\), where \( \mathbf{W}_Q \) and \( \mathbf{b}_Q \) represent the weights and biases. Therefore, this module enhances the model's generalization capability while maintaining computational efficiency.

\section{Experiments}

\subsection{Experiment Setup}
\subsubsection{Datasets}
Following the evaluation protocol of recent state-of-the-art DGSS methods, we conduct extensive experiments on various wide-ranging segmentation datasets. 
These include synthetic datasets (\emph{e.g.}, GTA5~\cite{Richter2016PlayingFD} and SYNTHIA~\cite{Ros2016TheSD}) as well as real-world datasets (\emph{e.g.}, Cityscapes~\cite{Cordts2016TheCD}, BDD-100K~\cite{Yu2020BDD100KAD}, Mapillary~\cite{Neuhold2017TheMV}, and ACDC~\cite{Sakaridis2021ACDCTA}).

\begin{table}[t]
    \centering
    \renewcommand\arraystretch{1.5}
    \setlength\tabcolsep{7pt}
    \resizebox{\columnwidth}{!}{ 
        \begin{tabular}{l|cccc}
            \toprule
            \textbf{DGSS Methods} & \textbf{Citys} & \textbf{BDD} & \textbf{Map} &\textbf{Avg.} \\
            \midrule
            \multicolumn{4}{l}{\textit{ResNet-based:}} \\
            ISW~\textsuperscript{\scriptsize\cite{choi2021robustnet}} & 36.58 & 35.20 & 40.33 &37.37\\
            GTR~\textsuperscript{\scriptsize\cite{Peng2021GlobalAL}} & 37.53 & 33.75 & 34.52 & 35.27 \\
            SHADE\textsuperscript{\scriptsize\cite{zhao2022style}} & 44.65 & 39.28 & 43.34 & 42.42 \\
            SAW~\textsuperscript{\scriptsize\cite{peng2022semantic}} & 39.75 & 37.34 & 41.86 & 39.65\\
            WildNet~\textsuperscript{\scriptsize\cite{lee2022wildnet}} & 44.62 & 38.42 & 46.09 & 43.04 \\
            SPC~\textsuperscript{\scriptsize\cite{huang2023spc}} & 44.10 & 40.46 & 45.51 &43.36  \\
            BlindNet~\textsuperscript{\scriptsize\cite{ahn2024styleblind}} & 45.72 & 41.32 & 47.08&44.71  \\
            \midrule
            \multicolumn{4}{l}{\textit{Transformer-based:}} \\
            CMFormer~\textsuperscript{\scriptsize\cite{cmf2024learning}} & 55.31 & 49.91 & 60.09 & 55.10 \\
            AdaptFormer~\textsuperscript{\scriptsize\cite{chen2022adaptformer}} &64.90 &59.00 &64.20 &62.70 \\
            \midrule
            \multicolumn{4}{l}{\textit{VFM-based:}} \\
            DIDEX\textsuperscript{\scriptsize\cite{niemeijer2024generalization}} & 62.00 & 54.30 & 63.00  & 59.77 \\
            Rein\textsuperscript{\scriptsize\cite{wei2024stronger}}& 66.40 & 60.40 & 66.10 &64.30 \\
            SET\textsuperscript{\scriptsize\cite{yi2024learning}} & 68.06 & 61.64 & 67.68 & 65.79\\
            FADA~\textsuperscript{\scriptsize\cite{bi2024learning}} & \underline{68.23} & 61.94 & \underline{68.09} & 66.09 \\
          DRF\textsuperscript{\scriptsize\cite{zhao2025fishertune}}  & 68.20 & \underline{63.30} & 68.00 & \underline{66.50}\\
            \textbf{Ours} (MGFC) & \underline{\textbf{69.65}} & \underline{\textbf{63.37}} & \underline{\textbf{69.47}} & \underline{\textbf{67.50}} \\
            \bottomrule
        \end{tabular}
    }
    \caption{Performance comparison between VFM-based methods and other DGSS approaches under the setting of  \{GTA5\} $\rightarrow$ \{Cityscapes, BDD-100K, Mapillary\}.}
    \vspace{-5pt}
    \label{tab:comparison_methods}
\end{table}

\subsubsection{Implementation Details}
By default we adopt frozen DINOv2~\cite{oquab2023dinov2} as the backbone, while also evaluate other VFMs, such as CLIP~\cite{radford2021clip}, SAM~\cite{kirillov2023sam}, MAE~\cite{he2022mae} and EVA02~\cite{fang2023eva02,fang2023eva}.Following prior works ~\cite{wei2024stronger,bi2024learning}, we employ Mask2Former~\cite{Cheng2021MaskedattentionMT} as the decoder with identical loss configurations.
We implement the model with a batch size of 4 on a single NVIDIA RTX3090 (24 GB).
We optimize using AdamW~\cite{kingma2015adam} with a weight decay of $0.05$. 
While keeping the backbone frozen, we train only the proposed modules and the decoder, with a learning rate of \(1 \times 10^{-4}\) applied to both.
To ensure efficient training, we train the network for 40K iterations and crop input images to a resolution of \(512 \times 512\). 
We evaluate segmentation performance using the PASCAL VOC Intersection over Union (IoU) metric~\cite{Everingham2014ThePV}. 
Mean IoU (mIoU) is computed as the average IoU across all categories.

\subsection{Comparison with State-of-the-Arts}
\begin{figure*}[t]
  \centering
\includegraphics[width=1.00\textwidth]{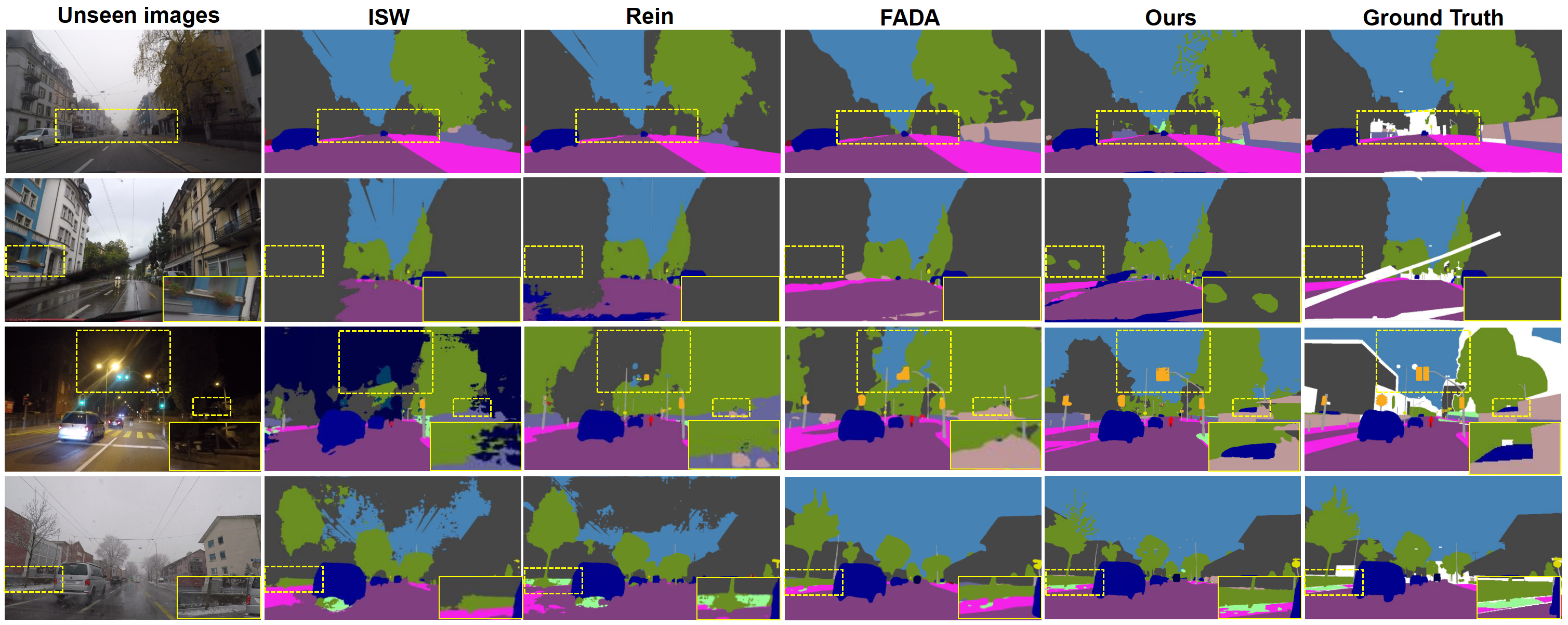}
   \caption{Qualitative results under adverse weather (From top to bottom: Fog, Rain, Night, Snow). The model is trained on the GTA5 and then generalized to the ACDC dataset.}
   \label{fig:acdcresult}
   \vspace{-6pt}
\end{figure*}

Tab.~\ref{tab:comparison_methods} presents a comprehensive comparison, where our method consistently outperforms existing DGSS techniques, including those based on both ResNet-based and Transformer-based 
\begin{table}[t]
\centering
\footnotesize
\setlength\tabcolsep{3pt}
\renewcommand\arraystretch{1.4}
\resizebox{\columnwidth}{!}{
\begin{tabular}{l|l|c|cccc}
\toprule
\textbf{VFM} & 
\textbf{Fine-tune Method} & 
\textbf{Trainable} & \textbf{Citys} & \textbf{BDD} & \textbf{Map} & \textbf{Avg.}\\
\midrule
\multirow{7}{*}{EVA02}
& Full        & 304.24M & 62.1 & 56.2 & 64.6 &60.9 \\
& Frozen      & 0.00M   & 56.5 & 53.6 & 58.6 &56.2 \\
& Rein\textsuperscript{\scriptsize\cite{wei2024stronger}}        & 2.99M  & 65.3 & 60.5 & 64.9 &63.6 \\
& SET\textsuperscript{\scriptsize\cite{yi2024learning}}          & 6.13M  & 66.4 & 61.8 & 65.6 &64.6 \\
& FADA\textsuperscript{\scriptsize\cite{bi2024learning}}         & 11.65M & \underline{66.7} & \underline{61.9} & \underline{66.1} &\underline{64.9} \\
& DRF\textsuperscript{\scriptsize\cite{zhao2025fishertune}} & 15.21M & 65.8 & 61.5 & 66.0 &64.4 \\
& MGFC        & 12.44M & \underline{\textbf{67.8}} & \underline{\textbf{62.9}} & \underline{\textbf{67.3}} &\underline{\textbf{66.0}}\\
\midrule
\multirow{7}{*}{SAM}
& Full        & 632.18M & 57.6 & 51.7 & 61.5 &56.9 \\
& Frozen      & 0.00M   & 57.0 & 47.1 & 58.4 &54.2 \\
& Rein\textsuperscript{\scriptsize\cite{wei2024stronger}}        & 4.51M  & 59.6 & 52.0 & 62.1  &57.9\\
& SET\textsuperscript{\scriptsize\cite{yi2024learning}}          & 9.21M  & 60.7 & 52.8 & 63.2 &58.9 \\
& FADA\textsuperscript{\scriptsize\cite{bi2024learning}}         & 16.59M & \underline{61.0} & 53.2 & 63.4 &59.2 \\
& DRF\textsuperscript{\scriptsize\cite{zhao2025fishertune}} & 15.21M & 60.9 &\underline{54.4}& \underline{63.9} &\underline{59.7} \\
& MGFC        & 17.31M & \underline{\textbf{62.0}} & \underline{\textbf{54.5}} & \underline{\textbf{64.6}} & \underline{\textbf{60.4}} \\
\midrule
\multirow{6}{*}{MAE}
& Full        & 304.20M & 53.7 & 50.8 & 58.1 &54.2\\
& Frozen      & 0.00M   & 43.3 & 37.8 & 48.0 &43.0 \\
& Rein\textsuperscript{\scriptsize\cite{wei2024stronger}}        & 2.99M  & 55.0 & 49.3 & 58.6 & 54.3\\
& SET\textsuperscript{\scriptsize\cite{yi2024learning}}          & 6.13M  & 56.2 & 51.0 & \underline{60.2} &55.8 \\
& DRF\textsuperscript{\scriptsize\cite{zhao2025fishertune}} & 15.21M & \underline{56.6} & \underline{51.9} & 59.7 &\underline{56.1} \\
& MGFC        & 12.44M & \underline{\textbf{57.3}} & \underline{\textbf{52.8}} & \underline{\textbf{61.3}}& \underline{\textbf{57.1}} \\
\midrule
\multirow{7}{*}{CLIP}
& Full        & 304.15M & 51.3 & 47.6 & 54.3 &51.1\\
& Frozen      & 0.00M   & 53.7 & 48.7 & 55.0 &52.4\\
& Rein\textsuperscript{\scriptsize\cite{wei2024stronger}}        & 2.99M  & 57.1 & 54.7 & 60.5 &57.4 \\
& SET\textsuperscript{\scriptsize\cite{yi2024learning}}          & 6.13M  & 58.2 & 55.3 & 61.4  &58.3\\
& FADA\textsuperscript{\scriptsize\cite{bi2024learning}}         & 11.65M & 58.7 & 55.8 & \underline{62.1} & 58.9\\
& DRF\textsuperscript{\scriptsize\cite{zhao2025fishertune}} & 15.21M & \underline{59.2} & \underline{\textbf{57.5}} & 61.0  &\underline{59.2}\\
& MGFC        & 12.44M & \underline{\textbf{60.2}} & \underline{57.4} & \underline{\textbf{63.2}} & \underline{\textbf{60.3}}\\
\bottomrule
\end{tabular}
}
\caption{
Performance comparison with different VFM configurations under the setting of  \{GTA5\} $\rightarrow$ \{Cityscapes, BDD-100K, Mapillary\}.}
\vspace{-8pt}
\label{tab:vfm}
\end{table}
approaches. In addition, Tab.~\ref{tab:vfm} presents further evaluations of our method using various VFMs to validate its generalizability and effectiveness. This superior performance is attributed to the combination of our vision foundation model backbone and the proposed multi-granularity feature calibration strategy, which effectively aligns feature representations across different granularities.
Furthermore, qualitative results presented in Fig.~\ref{result-city} highlight the practical advantages of our approach in semantic segmentation tasks. 
To further verify the effectiveness, we extend the application in real-world adverse scenarios. As shown in Tab.~\ref{tab:performance_comparison4} and Fig.~\ref{fig:acdcresult}, our method shows superior performance compared to the other approaches under the adverse weather. 
For example, in the highlighted regions, our method accurately segments fine structures such as poles, small traffic signs, and occluded objects. Notably, we visualize the manually annotated ground truth alongside the predictions and observe that some regions are overlooked in the annotations. In the first row of Fig.~\ref{fig:acdcresult}, we successfully identifies the white region that is missing in the ground truth. Similarly, in the rainy image, our model correctly segments green vegetation that was not labeled by human annotators.

\begin{table}[t]
    \centering
    \renewcommand\arraystretch{1.5}
    \setlength\tabcolsep{8pt}
    \resizebox{\columnwidth}{!}{ 
        \begin{tabular}{l|cccc}
            \toprule
            \textbf{DGSS Methods} & \textbf{Fog} & \textbf{Rain} & \textbf{Snow} & \textbf{Night}  \\
            \midrule
            \multicolumn{5}{l}{\textit{ResNet-based:}} \\
            IBN-Net~\textsuperscript{\scriptsize\cite{Pan2018TwoAO}} & 63.8 & 50.4 & 49.6 & 21.2 \\
            Iternorm~\textsuperscript{\scriptsize\cite{huang2019iterative}} & 63.3 & 50.1 & 49.9 & 23.8 \\
            IW~\textsuperscript{\scriptsize\cite{pan2019switchable}} & 62.4 & 52.4 & 47.6 & 21.8 \\
            ISW~\textsuperscript{\scriptsize\cite{choi2021robustnet}} & 64.3 & 56.0 & 49.8 & 24.3 \\
            \midrule
            \multicolumn{5}{l}{\textit{Transformer-based:}} \\
            HGFormer~\textsuperscript{\scriptsize\cite{ding2023hgformer}} & 69.9 & 72.0 & 68.6 & 52.7 \\
            CMFormer~\textsuperscript{\scriptsize\cite{cmf2024learning}} & 77.8 & 67.6 & 64.3 & 33.7 \\
            \midrule
            \multicolumn{5}{l}{\textit{VFM-based:}} \\
            Rein~\textsuperscript{\scriptsize\cite{wei2024stronger}} & 79.48 & 72.45 & 70.57 & 55.92 \\
            SET~\textsuperscript{\scriptsize\cite{yi2024learning}} & 80.06 & 74.80 & \underline{73.69} & 57.29 \\
            FADA\textsuperscript{\scriptsize\cite{bi2024learning}} & \underline{80.20} & \underline{75.00} & 73.50 & \underline{57.40} \\
            \textbf{Ours} (MGFC) & \underline{\textbf{81.83}} & \underline{\textbf{76.06}} & \underline{\textbf{74.93}} & \underline{\textbf{57.91}} \\
            \bottomrule
        \end{tabular}
    }
    \caption{Performance comparison between VFM-based methods and other DGSS approaches under the setting of  \{Cityscapes\} $\rightarrow$ \{ACDC\}.}
    \vspace{-8pt}
    \label{tab:performance_comparison4}
\end{table}

\begin{table}[t]
    \centering
    \renewcommand\arraystretch{1.5}
    \setlength\tabcolsep{8pt}
    \resizebox{0.88\columnwidth}{!}{
        \begin{tabular}{ccc|cccc}
            \toprule
            
            \textbf{CGT} & \textbf{MGT} & \textbf{FGT} & \textbf{Citys} & \textbf{BDD} & \textbf{Map}& \textbf{Avg.} \\
            \midrule
            \checkmark & - & - & 67.04 & 60.83 & 66.76 & 64.88\\ 
            - & \checkmark  & - & 67.34 & 61.12 & 67.29  & 65.25\\ 
            - & - & \checkmark   & 67.82 & 61.53 & 67.41& 65.59\\ 
            \midrule
            \checkmark & \checkmark & -  & 68.29 & 61.84 & 67.90 & 66.01\\ 
            \checkmark & - & \checkmark  & 68.52 & 62.11 & 68.27  & 66.30 \\ 
            - & \checkmark & \checkmark   & 68.90 & 62.54 & 68.68 & 66.71 \\ 
            \midrule
            \checkmark & \checkmark & \checkmark  & \textbf{69.65} & \textbf{63.37} & \textbf{69.47} & \textbf{67.51}  \\ 
            \bottomrule
        \end{tabular}
    }
    \caption{Ablation on the components of Tuner Modules.}
    \vspace{-6pt}
    \label{tab:frequency_table}
\end{table}

\subsection{Ablation Study}

\subsubsection{\textbf{Effect of Multi-granularity Tuners}}
We conduct ablation studies on the multi-granularity tuner by evaluating the individual and combined effects of each tuner branch. As shown in Tab.~\ref{tab:frequency_table}, removing any of the tuners leads to a noticeable performance degradation compared with the fully configured network. This demonstrates that the combination of different granularity levels contributes jointly to the overall model performance. Each tuner captures complementary information at different levels. The observed performance degradation under each ablation setting validates the necessity of integrating tuners at multiple granularity levels.

\subsubsection{\textbf{Analysis of Clustering Strategy}}
In the CGT, we apply a clustering-based strategy to segment spatial features into semantically meaningful regions. 
Specifically, we adopt DBSCAN clustering to partition feature maps into different clusters, and then apply instance normalization within each region to perform localized style alignment across domains.
To evaluate the effect of different clustering approaches, we also adopt and test the K-Means strategy.
As shown in Tab.~\ref{tab:ablation_dbscan}, CGT consistently achieves better performance under the DBSCAN. Moreover, it maintains strong performance across different parameter settings, demonstrating the adaptability of our framework to different clustering strategies.

\section{Conclusion}
This paper studied the problem of domain generalized semantic segmentation via VFMs, aiming to enable pixel-wise prediction on unseen target domains without any target supervision. To address the challenge of domain shifts across appearance, structure, and texture, the Multi-Granularity Feature Calibration framework is proposed. 
Distinct from other VFM-based methods, our approach adopts a multi-granularity perspective. By hierarchically aligning features at coarse, medium, and fine levels, MGFC enhances global semantics, category-level discrimination, and fine-grained spatial details. Additionally, a query fusion module is designed to integrate multi-level contextual cues and guide effective cross-granularity representation learning. 
Extensive experiments under both normal and adverse conditions show that MGFC surpasses state-of-the-art methods, validating the effectiveness of granularity-aware feature calibration for domain generalization in semantic segmentation.  

\begin{table}[!t]
    \centering
    \renewcommand\arraystretch{1.5}
    \setlength\tabcolsep{8pt}
    \resizebox{0.88\columnwidth}{!}{ 
        \begin{tabular}{l|c|ccc}
            \toprule
            \textbf{Cluster} & \textbf{Settings} & \textbf{Citys} & \textbf{BDD} & \textbf{Map} \\
            \midrule
            \multirow{3}{*}{K-Means}
            & $k=3$ & 69.02 & 62.60 & 68.83 \\
            & $k=5$ & 69.14 & 62.71 & 68.97 \\
            & $k=7$ & 69.09 & 62.63 & 68.92 \\
            \midrule
            \multirow{5}{*}{DBSCAN}
            & $\varepsilon=15$, minPts=3   & 69.13  &62.87    &69.18 \\
            & $\varepsilon=15$, minPts=4   & 69.27  &62.92    &69.21\\
            & $\varepsilon=20$, minPts=3   & 69.58  &63.23    &69.43\\     
            & $\varepsilon=20$, minPts=4   & \textbf{69.65}   &\textbf{63.37}  & \textbf{69.47}\\
            & $\varepsilon=20$, minPts=5    & 69.47   &62.98    &69.24 \\
            \bottomrule
        \end{tabular}
    }
    \caption{Ablation study on clustering strategies in CGT.}
    \label{tab:ablation_dbscan}
    \vspace{-6pt}
\end{table}

\bibliography{aaai2026}

\end{document}